\documentclass[11pt,letterpaper]{article}
\usepackage{naaclhlt2010}
\usepackage[numbers]{natbib}
\usepackage{times}
\usepackage{latexsym}
\usepackage{amsmath}\usepackage{fancybox}
\usepackage{xspace,graphicx,url}
\newcommand{\omt}[1]{}

\newcommand{\ex}[1]{``{\it #1}''}

\newcommand{\simped}[2]{\ex{#1} $\to$ \ex{#2}}

\newcommand{\edit}[1]{e_{k,#1}}

\newcommand{\versions}{\vec{d}_k} 
\newcommand{\revvec}{\vec{r}_k} 
\newcommand{\rev}[1]{r^{#1}_k} 
\newcommand{\cmt}[1]{c^{#1}_k} 

\newcommand{\edits}{e_k}

\newcommand{\OP}{\Omega}
\newcommand{\CW}{C\xspace}
\newcommand{\SW}{S\xspace}
\newcommand{\op}[1]{#1}
\newcommand{\fix}{\op{fix}\xspace}
\newcommand{\fixes}{\op{fixes}\xspace}
\newcommand{\simp}{\op{simplify}\xspace}
\newcommand{\noop}{\op{no-op}\xspace}
\newcommand{\spam}{\op{spam}\xspace}

\newcommand{\condp}[2]{{P}(#1 \mid #2)}
\newcommand{\hatp}[2]{{\widehat{P}}(#1 \mid #2)}
\newcommand{\sfreq}[2]{f_{#1}(#2)}

\newcommand{\revset}{TRev\xspace}
\newcommand{\revfunc}{R\xspace}

\newcommand{\SEED}{Seed\xspace}
\newcommand{\RANK}[1]{\mbox{RANK}(#1)\xspace}

\newcommand{\filtersimpl}{{\sc Simpl}\xspace}

\newcommand{\splist}{{\sc SpList}\xspace}
\newcommand{\randombaseline}{{\sc Random}\xspace}
\newcommand{\frequentbaseline}{{\sc Frequent}\xspace}

\newcommand{\sew}{{SimpleEW}\xspace}
\newcommand{\cew}{{ComplexEW}\xspace}

\title{For the sake of simplicity: \\ Unsupervised extraction of lexical simplifications from Wikipedia}

\author{Mark Yatskar,  Bo Pang,  Cristian Danescu-Niculescu-Mizil and Lillian Lee \\ my89@cornell.edu, bopang@yahoo-inc.com, cristian@cs.cornell.edu, llee@cs.cornell.edu}
\begin{document}
\maketitle
\begin{abstract}
\vspace*{-.15in}
We report on work in progress on extracting 
lexical simplifications (e.g., \simped{collaborate}{work together}),
focusing on 
 utilizing
edit histories in Simple English Wikipedia for this task. 
We consider two main approaches:
(1) deriving simplification probabilities via an edit model 
that accounts for a mixture of different operations, 
and (2) using metadata to focus on edits that are more likely to
be simplification operations.
We 
find our methods to
outperform a reasonable baseline and 
 yield many high-quality
lexical simplifications not included in 
an independently-created 
manually prepared list.  

{\bf Published at}: NAACL 2010 (short paper)

\end{abstract}
\vspace*{-.25in}

\section{Introduction}

{\it Nothing is more simple than greatness; indeed, to be simple is to be great.} \mbox{---Emerson}, {\it Literary Ethics}

\smallskip

Style is an important aspect of information presentation; indeed, 
different contexts call
for different styles. 
Here, we consider an important
dimension of style, namely, {\em simplicity}.
Systems that can rewrite text
into simpler versions promise to make information available to a
broader audience, such as non-native speakers, children, laypeople,
and so on.

One major effort
to produce such text is
the Simple English Wikipedia (henceforth \sew)\footnote{{\url{http://simple.wikipedia.org}}},
a sort of spin-off of the well-known English Wikipedia (henceforth \cew)
where human editors
enforce simplicity of language through rewriting.
The crux of our proposal  
is to 
learn lexical simplifications from \sew edit histories, thus leveraging the efforts of the 18K pseudonymous individuals who work on \sew.
 Importantly,  not all the changes on \sew are simplifications; 
we thus also make use of \cew edits to filter out non-simplifications.

\paragraph{Related work and related problems}
Previous 
work
usually  involves
general syntactic-level transformation rules
 \citep{Chandrasekar+Srinivas:97a,Siddharthan+Nenkova+McKeown:04a,Vickrey+Koller:08a}.\footnote{
One exception \citep{Klebanov+Knight+Marcu:04a}
 changes verb tense and 
 replaces pronouns.
%
 Other lexical-level work  focuses on 
medical text \citep{Elhadad+Sutaria:07a,Deleger+Zweigenbaum:09a},
 or uses frequency-filtered WordNet synonyms \citep{Devlin+Tait:98a}.
 }  In contrast, we explore data-driven methods to learn 
{\em lexical simplifications} (e.g., \simped{collaborate}{work together}), which are highly specific to the lexical items involved and thus 
cannot be captured by a few general rules.

Simplification is strongly related to  but distinct from paraphrasing and machine translation (MT).  
While it can be considered a directional form of the former, it differs in spirit because 
simplification
must trade off meaning preservation (central to paraphrasing) against complexity reduction (not 
a consideration in paraphrasing).  Simplification can also
be considered to be a form of MT in which the two
``languages" in question are 
highly related.  However, 
note that \cew and \sew do not 
together
constitute a clean parallel corpus, but 
rather an extremely noisy comparable corpus.  For example,  
Complex/Simple same-topic document pairs are often written completely independently of each other, and 
even when it is possible to get good sentence alignments between them, the sentence pairs may reflect operations other than simplification, such as corrections, 
additions, or
edit spam.

Our work joins others in using Wikipedia revisions to learn interesting types of directional lexical relations,
e.g,
``eggcorns"\footnote{
A type of lexical corruption, e.g., ``acorn"$\to$``eggcorn".} \citep{Nelken+Yamangil:08a} and entailments \citep{Shnarch+Barak+Dagan:09a}.

\section{Method}
\label{sec:method}
As mentioned
 above, 
 a key idea in our work is 
to utilize \sew edits.
The 
primary difficulty in working with 
these modifications
is that 
they
include
not only
 simplifications 
but also
edits that
serve
other
functions, such as spam removal or correction of grammar or factual content
(``fixes'').
We describe two main approaches to 
this problem:
a probabilistic model
that 
captures this mixture of different edit operations (\S \ref{sec:method:edit}),
and
the use of
{\em metadata}
to filter out undesirable revisions (\S \ref{sec:method:metadata}).

\subsection{Edit model}
\label{sec:method:edit}

We say that the $k^{th}$  article in 
a
Wikipedia corresponds to (among other things) a title or {\em topic} (e.g.,
``Cat") and a sequence $\versions$ of article versions caused by successive
edits.   For a
given lexical item or phrase $A$, we write  $A \in \versions$ if there is any version in $\versions$ that contains
$A$.   From each $\versions$ we extract a collection 
$\edits=(\edit{1}, \edit{2}, \ldots, \edit{n_k})$ of {\em
lexical edit instances}, {\em repeats allowed},  where  $\edit{i} = A \to a$ means that phrase $A$ in one version 
was changed to $a$ in the next, $A \neq a$; 
e.g., 
\simped{stands for}{is the same as}.  
(We defer detailed  description of how we extract lexical edit instances from data to \S \ref{sec:data}.)
We denote the collection of 
$\versions$
in \cew and \sew 
as $\CW$ and $\SW$, respectively.

There are at least
four possible edit operations:
{\em \fix} ($o_1$), {\em \simp} ($o_2$), {\em \noop} ($o_3$), or {\em \spam} ($o_4$).
However, for this initial work we assume $P(o_4)=0$.\footnote
{Spam/vandalism detection
is 
a direction 
for future work.}

Let $\condp{o_i}{A}$ be the probability that $o_i$ is applied to $A$, and $\condp{a}{A, o_i}$ be the probability of $A \to a$ given that the operation is $o_i$.
The key quantities of interest are  \mbox{$\condp{o_2}{A}$} in $\SW$, which is 
the probability that $A$ should be simplified, and  $\condp{a}{A, o_2}$, 
which yields proper simplifications of $A$.  We start with an equation that models 
the probability that a phrase $A$ is edited into $a$:
{
\begin{equation}
\label{eq:mix}
\condp{a}{A} = \sum_{o_i \in \OP}{\condp{o_i}{A} \condp{a}{A, o_i}},
\end{equation}
}
where $\OP$ is the set of edit operations.
This involves the desired parameters, which
we solve
 for by estimating the 
others
from data, as described next.

\paragraph{Estimation}
Note that {\small $\condp{a}{A,o_3} = 0 \mbox{ if } A \ne a$}.
Thus, if we have estimates for  $o_1$-related probabilities,
we can derive $o_2$-related probabilities via Equation \ref{eq:mix}.
To begin with, we make the working assumption that  occurrences of simplification in \cew are negligible in comparison to fixes.  
Since we are also currently ignoring edit spam,  we thus assume
that only $o_{1}$ edits occur in \cew.\footnote{
This assumption 
also
provides useful constraints to EM, which we plan to apply in the  future, by reducing the number of parameter settings yielding the same likelihood.}

Let $\sfreq{\CW}{A}$ be the 
fraction
of $\versions$ in $\CW$ containing $A$
in which $A$ is modified:
\centerline{
$\sfreq{\CW}{A} = \frac{|\{ \versions \in \CW \mid  \exists a,i \mbox{ such that }  \edit{i} = A \to a \}|}{|\{\versions \in \CW \mid A \in \versions \}|}.$
}
We similarly define $\sfreq{\SW}{A}$ on $\versions$ in $\SW$.
Note that we count topics (version sequences), not individual versions: if $A$ appears at some point and is not edited until 
50 revisions later, we should {\em not}  conclude that $A$ is unlikely to be rewritten; for example, the intervening revisions could all be minor additions, or part of an edit war.

If we assume that the probability of 
any particular
 \fix operation 
 being
applied
in \sew 
is proportional to that
in \cew
--- e.g.,  the  \sew \fix rate might be dampened  because  already-edited \cew articles are copied over --- we have\footnote{Throughout, ``hats''  denote estimates.}
{
$$\hatp{o_1}{A} = \alpha\sfreq{\CW}{A}$$
}
where $0 \le \alpha \le 1$.
Note that in \sew,
{\small
$$\condp{o_1 \vee o_2}{A} = \condp{o_1}{A} + \condp{o_2}{A},$$
}
where $\condp{o_1 \vee o_2}{A}$ is the probability 
that $A$ is changed to 
a different word
in \sew,
which we estimate as $\hatp{o_1 \vee o_2}{A} = \sfreq{\SW}{A}$.  We then
set
\begin{center}\boxed{\mathbf{\hatp{o_2}{A} = \max\left(0,\sfreq{\SW}{A} - \alpha\sfreq{\CW}{A}\right)}.}\end{center}

Next, under our working assumption, we estimate the probability of $A$ being changed to $a$ as a fix by the proportion 
of 
\cew
edit instances that rewrite $A$ to $a$:
{\small
$$\hatp{a}{A, o_1} = \frac{|\{  (k,i) \mbox{ pairs}  \mid  \edit{i}=A \to a   \wedge \versions\in C             \}|}
{ \sum_{a'} |\{  (k,i) \mbox{ pairs}  \mid  \edit{i}=A \to a'   \wedge \versions\in C             \}|     }.$$
}

A natural estimate  for the conditional probability of $A$ being rewritten to $a$ under any operation type is based on observations of $A \to a$ 
in 
\sew, since that is the corpus wherein both operations are assumed to occur:
{\small
$$
\hatp{a}{A} = \frac{  |\{  (k,i) \mbox{ pairs}  \mid  \edit{i}=A \to a    \wedge \versions \in \SW          \}| }
{ \sum_{a'} |\{  (k,i) \mbox{ pairs}  \mid  \edit{i}=A \to a'          \wedge \versions \in \SW        \}|   }
.$$
}
Thus,
from (\ref{eq:mix}) we get that
for $A \ne a$:
{\small \begin{center}
\boxed{\mathbf{\hatp{a}{A, o_2} = \frac{\hatp{a}{A} - \hatp{o_1}{A} \hatp{a}{A, o_1}}{\hatp{o_2}{A}}.}}
\end{center}
}

\subsection{Metadata-based methods}
\label{sec:method:metadata}
Wiki editors
have the option of 
associating a comment with each revision,  
and such comments sometimes indicate
the intent of the revision.
We therefore sought to use comments to identify ``trusted" revisions wherein the extracted lexical edit instances (see \S \ref{sec:data}) would be likely to be simplifications.

Let $\revvec = (\rev{1}, \ldots, \rev{i}, \ldots)$ be the sequence of revisions for the $k^{th}$ article in
\sew, 
where $\rev{i}$ is the set of lexical edit instances ($A \to a$) extracted from the $i^{th}$
modification of the document.  
Let $\cmt{i}$ be the comment that accompanies 
$\rev{i}$,
and conversely,
let $\revfunc(Set) = \{\rev{i} | \cmt{i} \in Set \}$.

We start with a seed set of trusted comments,  $\SEED$.  
To initialize it, we manually inspected  a small sample of the 700K$+$ \sew revisions that bear comments, and found that comments containing a word matching the regular expression
  {\sf*simpl*} (e.g, ``simplify") seem promising. We thus set 
$\SEED := \{{\sf *simpl*}\}$ (abusing notation). 

\paragraph{The \filtersimpl method} Given a set of trusted revisions
$\revset$
(in our case
$\revset = \revfunc(\SEED)$),
we score each $A \to a \in \revset$ 
by the point-wise mutual information (PMI) between $A$ and $a$.\footnote{PMI seemed to outperform raw frequency and conditional probability.}  We write $\RANK{\revset}$ to denote the PMI-based ranking of $A \to a \in \revset$, and use \filtersimpl to denote our most basic ranking method,  $\RANK{\revfunc(\SEED)}$.

\paragraph{Two ideas for bootstrapping}
We also
considered bootstrapping as a way to
be able to 
utilize
revisions whose comments are not in the initial $\SEED$ set.

Our first idea was
to iteratively expand the set of
 {trusted} comments 
to include those that 
most often accompany 
already highly ranked simplifications. 
Unfortunately, 
our initial implementations
involved many parameters (upper and lower comment-frequency thresholds, number of highly ranked simplifications to consider, number of comments to add per iteration),
making it
relatively difficult to tune;
we
thus
 omit its results.

Our second idea was to iteratively expand the set of trusted revisions, adding those that contain already highly ranked simplifications.
While our initial implementation had fewer parameters than the method sketched above, it tended to terminate quickly, so that not many new simplifications were found;
so, again, we do not report results here.

An important direction for future work is to  differentially weight the edit instances within a revision, as opposed to placing equal trust in all of them; this could prevent our bootstrapping methods from giving common \fixes (e.g., \simped{a}{the}) high scores.

\section{Evaluation\footnote{Results at http://www.cs.cornell.edu/home/llee/data/simple}}

\subsection{Data}
\label{sec:data}

We obtained the revision histories of both
\sew (November 2009 snapshot) and \cew (January 2008
snapshot).
In total, 
$\sim$1.5M revisions for 81733 \sew articles were
processed (only 30\% involved textual changes).
For \cew, we processed
$\sim$16M  revisions for 19407 articles.

\paragraph{Extracting lexical edit instances.} 
For each article, we aligned sentences in each pair of adjacent versions
using tf-idf scores in a way similar to 
\citet{Nelken+Shieber:06a} 
(this produced satisfying results because revisions tended to represent small changes).
From the aligned sentence pairs, we obtained the aforementioned 
lexical edit instances
$A \to a$.
Since the focus of our study was not word alignment, 
we used a simple method that identified the 
longest differing segments (based on word boundaries) between each sentence, 
except that to prevent the extraction of entire (highly non-matching) sentences, 
we
 filtered out $A \to a$ pairs if either $A$ or $a$ contained more than five words.

\subsection{Comparison points}
\label{sec:compare}
 \paragraph{Baselines} \randombaseline returns lexical edit instances drawn uniformly at random from among those extracted from \sew.  \frequentbaseline returns the most frequent lexical edit instances extracted from \sew.

\paragraph{Dictionary of simplifications}

The \sew editor ``Spencerk" (Spencer Kelly) 
has assembled a list of simple words and simplifications using a combination of dictionaries and manual
 effort\footnote{http://www.spencerwaterbed.com/soft/simple/about.html}. 
He provides a list of 17,900 simple words --- words that do not need further simplification ---   and a list of 2000 transformation pairs. 
We did not use Spencerk's set as the gold standard
because many transformations we found to be reasonable were not
on his list. 
Instead, we
 measured our agreement with the list of transformations he assembled (\splist).
 
\subsection{Preliminary results}

The top 100 pairs from each system (edit model\footnote{
We only considered those $A$ such that  $freq(A\to*) > 1 \wedge freq(A) > 100$ on both \sew and \cew.  The final top 100 $A \to a$ pairs were those with $A$s with  the highest $\condp{o_2}{A}$.
We set $\alpha=1$. 
} 
 and \filtersimpl and
 the two baselines)
plus
100 randomly selected pairs from \splist
were mixed and
all
 presented in random order to 
three native English speakers and three non-native English speakers
(all non-authors).
Each pair was 
presented in random orientation (i.e., either as $A \to a$ or as $a \to A$),
and the labels included
``simpler", ``more complex", ``equal'', ``unrelated'', and 
``?'' (``hard to judge'').
The first two labels correspond to 
simplifications
for the orientations $A \to a$ and $a \to A$, respectively.
Collapsing the 5 labels into ``simplification'', ``not a simplification'', and 
``?'' yields 
reasonable agreement among the 3 native speakers ($\kappa = 0.69$;
75.3\% of the time all three 
agreed on the same label).
While we postulated that
non-native speakers\footnote{Native languages: Russian;  Russian; Russian and
Kazakh.} might be more sensitive to what was simpler,
we note that they disagreed more than the native speakers ($\kappa=0.49$) and reported having to 
consult 
a dictionary.
The 
native-speaker
majority label was used in our evaluations.

Here are the results;
``-x-y'' means that x and  y are the number of instances discarded from the precision calculation for having no majority label or majority label 
``?'', respectively:

\omt
{\small
\hspace{-.3in}\begin{tabular}{|r||l@{}|c@{}|c|} \hline
Method &Prec@100 &\# of pairs & right and $ \notin$   \splist \\ \hline
\splist    & 86 (-0-0)   &   & \\
Edit model & 77 (-0-1)  & 1079& 62\% \\ 
\filtersimpl &66 (-0-0)  &2970 & 71\% \\
\frequentbaseline    & 17 (-1-7)   & 2970   &  71\%\\
\randombaseline   &   17 (-1-4)  & 2970  & 94\%\\
\hline
\end{tabular}
}

{\small
\begin{center}
\begin{tabular}{|r||c|c|} \hline
Method &Prec@100 &\# of pairs  \\ \hline
\splist    & 86\% (-0-0)   &2000 \\
Edit model & 77\% (-0-1)  & 1079 \\ 
\filtersimpl &66\% (-0-0)  &2970 \\
\frequentbaseline    & 17\% (-1-7)   & -\\
\randombaseline   &   17\% (-1-4)  & -\\
\hline
\end{tabular}
\end{center}
}

Both baselines yielded very low precisions --- clearly not all (frequent) edits in \sew were simplifications.
Furthermore, the edit model yielded higher precision than \filtersimpl for the top 100 pairs.
(Note that we only examined one simplification per $A$ for those $A$
where $\hatp{o_2}{A}$ was well-defined; thus ``\# of pairs'' does not directly reflect the full potential
recall that either method can achieve.)
Both, however, produced many high-quality pairs
 (62\% and 71\% of the
{\em correct} pairs) {\em not} included in \splist.
We also found the pairs produced by these two systems to be 
complementary
 to each other.
We believe that these two approaches provide a good starting point
for further explorations.

Finally, some examples of simplifications found by our methods: \simped{stands for}{is the same as},
\simped{indigenous}{native}, \simped{permitted}{allowed}, 
\simped{concealed}{hidden}, \simped{collapsed}{fell down},
\simped{annually}{every year}.

\subsection{Future work}  Further evaluation could include comparison with machine-translation and paraphrasing algorithms.  It would be interesting to use our proposed estimates as initialization for EM-style iterative re-estimation. Another idea would be to estimate {\em simplification priors} based on 
a
model of inherent lexical complexity; some possible starting points are  number of syllables (which is used in various readability formulae) or word length.

\newcommand{\bibsnip}{\vspace*{-.08in}}

{\footnotesize 
\newcommand{\fname}[1]{.}
\noindent {\bf Acknowledgments}
We first wish to thank Ainur Yessenalina for initial investigations and helpful comments.  We
are also thankful to
R\fname{egina} Barzilay,
T\fname{om} Bruce,
C\fname{hris} Callison-Burch,
J\fname{ared} Cantwell,
M\fname{ark} Dredze, 
C\fname{ourtney} Napoles,
E\fname{vgeniy} Gabrilovich,
\&
the reviewers
for helpful comments; 
W. Arms and L. Walle for access to the Cornell Hadoop cluster;
J. Cantwell for access to computational resources;
R\fname{ebecca} Hwa
\&
A\fname{ndrew} Owens
for annotation software;
M\fname{organ} Ulinski
for 
preliminary explorations;
J\fname{ared} Cantwell, 
M\fname{yle} Ott,
J\fname{amie} Silverstein,
J\fname{ulia} Yatskar,
Y\fname{uri} Yatskar,
\&
A\fname{inur} Yessenalina
for annotations. 
Supported by NSF grant  IIS-0910664.

}
\newcommand{\supershrink}[1]{}

\end{document}